\documentclass{article}

    \PassOptionsToPackage{numbers, compress}{natbib}

\usepackage[preprint]{neurips_2022}

\usepackage[utf8]{inputenc} 
\usepackage[T1]{fontenc}    
\usepackage{hyperref}       
\usepackage{url}            
\usepackage{booktabs}       
\usepackage{amsfonts}       
\usepackage{nicefrac}       
\usepackage{microtype}      
\usepackage{xcolor}         
\usepackage{subcaption}
\usepackage{natbib}
\usepackage{comment}
\usepackage{hyperref}
\hypersetup{
    colorlinks=true,
    citecolor=black,
    linkcolor=red,
    urlcolor=cyan
}

\title{A Neural Scaling Law from Lottery Ticket Ensembling}

\author{%
  Ziming Liu \\ 
  MIT \& IAIFI \\
  \texttt{zmliu@mit.edu} \\
   \And
  Max Tegmark \\
  MIT \& IAIFI \\
  \texttt{tegmark@mit.edu} \\
}

\usepackage{graphicx}
\usepackage{dcolumn}
\usepackage{bm}
\usepackage{float}
\usepackage{subcaption}
\usepackage{multirow}
\usepackage{comment}
\usepackage{dsfont}
\usepackage{amsthm}
\usepackage{xcolor}
\usepackage{array}
\usepackage{eucal}
\usepackage{makecell}
\usepackage{mathtools}
\usepackage{makecell}

\usepackage{enumitem}
\setlist{leftmargin=10mm}

\newcommand{\mat}[1]{\mathbf{#1}}

\newcommand{\appropto}{\mathrel{\vcenter{
  \offinterlineskip\halign{\hfil$##$\cr
    \propto\cr\noalign{\kern2pt}\sim\cr\noalign{\kern-2pt}}}}}

\def\spose#1{\hbox to 0pt{#1\hss}}
\def\simlt{\mathrel{\spose{\lower 3pt\hbox{$\mathchar"218$}}
     \raise 2.0pt\hbox{$\mathchar"13C$}}}
\def\simgt{\mathrel{\spose{\lower 3pt\hbox{$\mathchar"218$}}
     \raise 2.0pt\hbox{$\mathchar"13E$}}}
\def\simpropto{\mathrel{\spose{\lower 3pt\hbox{$\mathchar"218$}}
     \raise 2.0pt\hbox{$\propto$}}}

\def\beq#1{\begin{equation}\label{#1}}
\def\eeq{\end{equation}}
\def\beqa#1{\begin{eqnarray}\label{#1}}
\def\eeqa{\end{eqnarray}}

\begin{document}

\maketitle

\begin{abstract}
    {\huge \textcolor{red}{Authors' note: the theory in this paper is questionable; we are trying our best to fix it. Empirical results still stand.}}
    
    Neural scaling laws (NSL) refer to the phenomenon where model performance improves with scale. Sharma \& Kaplan analyzed NSL using approximation theory and predict that MSE losses decay as $N^{-\alpha}$, $\alpha=4/d$, where $N$ is the number of model parameters, and $d$ is the intrinsic input dimension. Although their theory works well for some cases (e.g., ReLU networks), we surprisingly find that a simple 1D problem $y=x^2$ manifests a different scaling law ($\alpha=1$) from their predictions ($\alpha=4$). We opened the neural networks and found that the new scaling law originates from \textit{lottery ticket ensembling}: a wider network on average has more "lottery tickets", which are ensembled to reduce the variance of outputs. We support the ensembling mechanism by mechanistically interpreting single neural networks, as well as  studying them statistically. We attribute the $N^{-1}$ scaling law to the "central limit theorem" of lottery tickets. Finally, we discuss its potential implications for large language models and statistical physics-like theories of learning.   
\end{abstract}




\section{Introduction}

Neural scaling laws (NSL), the phenomenon where model performance improves as the model size scales up, has been widely observed in deep learning ~\cite{hestness2017deep, rosenfeld2019constructive, kaplan2020scaling, henighan2020scaling, gordon2021data, zhai2022scaling, hoffmann2022training,michaud2023quantization}. Typically the losses follow $\ell\propto N^{-\alpha}$ where $N$ is the number of model parameters, and $\alpha>0$ is the scaling exponent. Understanding the mechanisms of neural scaling laws is important both theoretically and empirically. Currently, two main theories of neural scaling laws are proposed: one is the \textit{approximation theory}~\cite{kaplan2020scaling, sharma2020neural,michaud2023precision}, claiming that the scaling law comes from regression on a data manifold. In particular, the scaling exponent is $\alpha=4/d$ where $d$ is the intrinsic input dimension~\cite{kaplan2020scaling,sharma2020neural} or maximum arity~\cite{michaud2023precision}. The other is the \textit{quanta theory}~\cite{michaud2023quantization}, which suggests that the scaling law comes from the hierarchy of subtasks, and the scaling exponent $\alpha$ depends on the heavy-tailedness of the subtask distribution. 

\begin{figure}[t]
    \centering
    \begin{subfigure}[]{0.23\textwidth}
\includegraphics[width=\linewidth, trim=0cm 0cm 0cm 0cm]{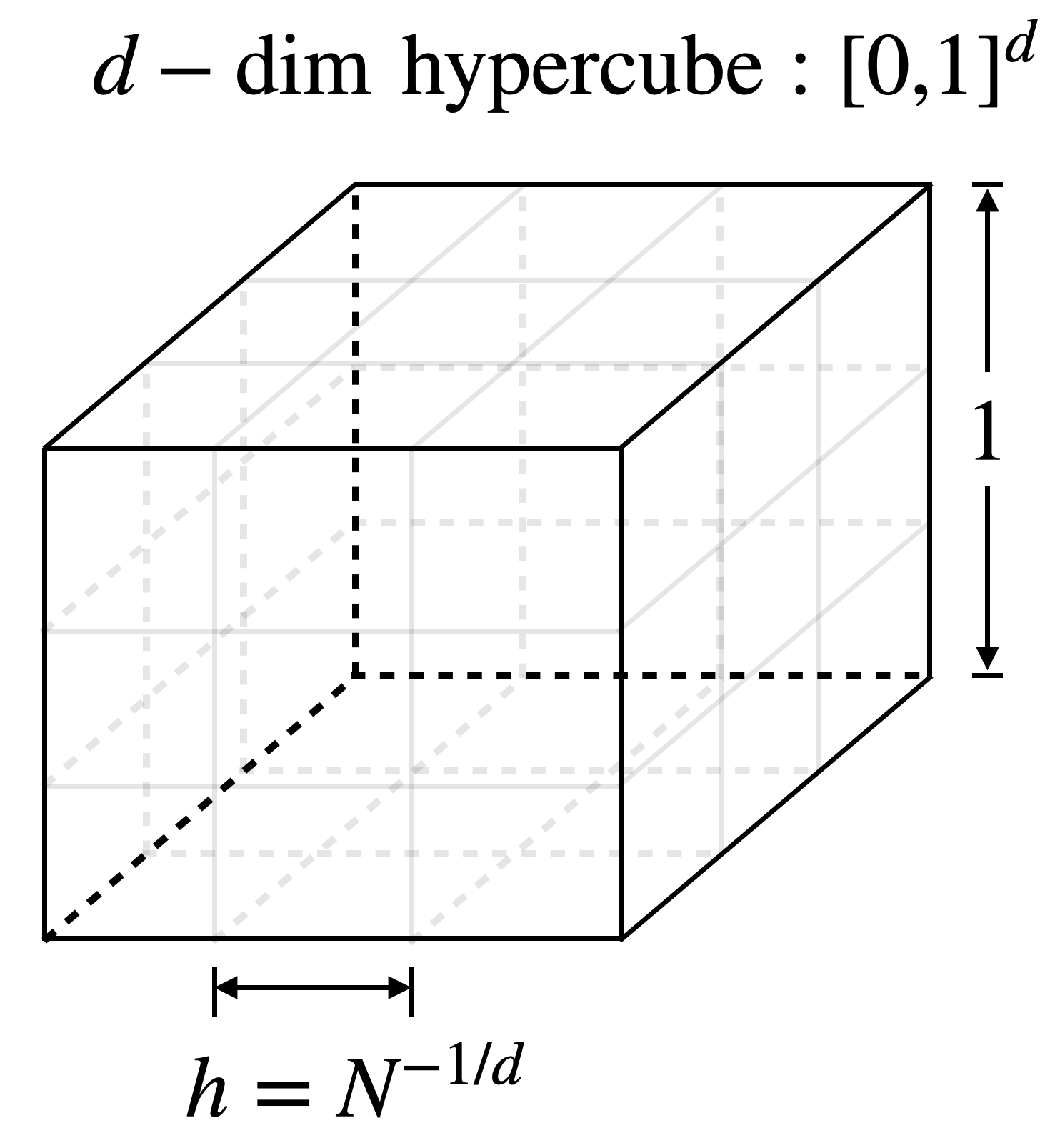}
    \caption{}
    \label{fig:hypercube}
    \end{subfigure}
    \begin{subfigure}[]{0.37\textwidth}
    \includegraphics[width=\linewidth, trim=0cm 0cm 0cm 0cm]{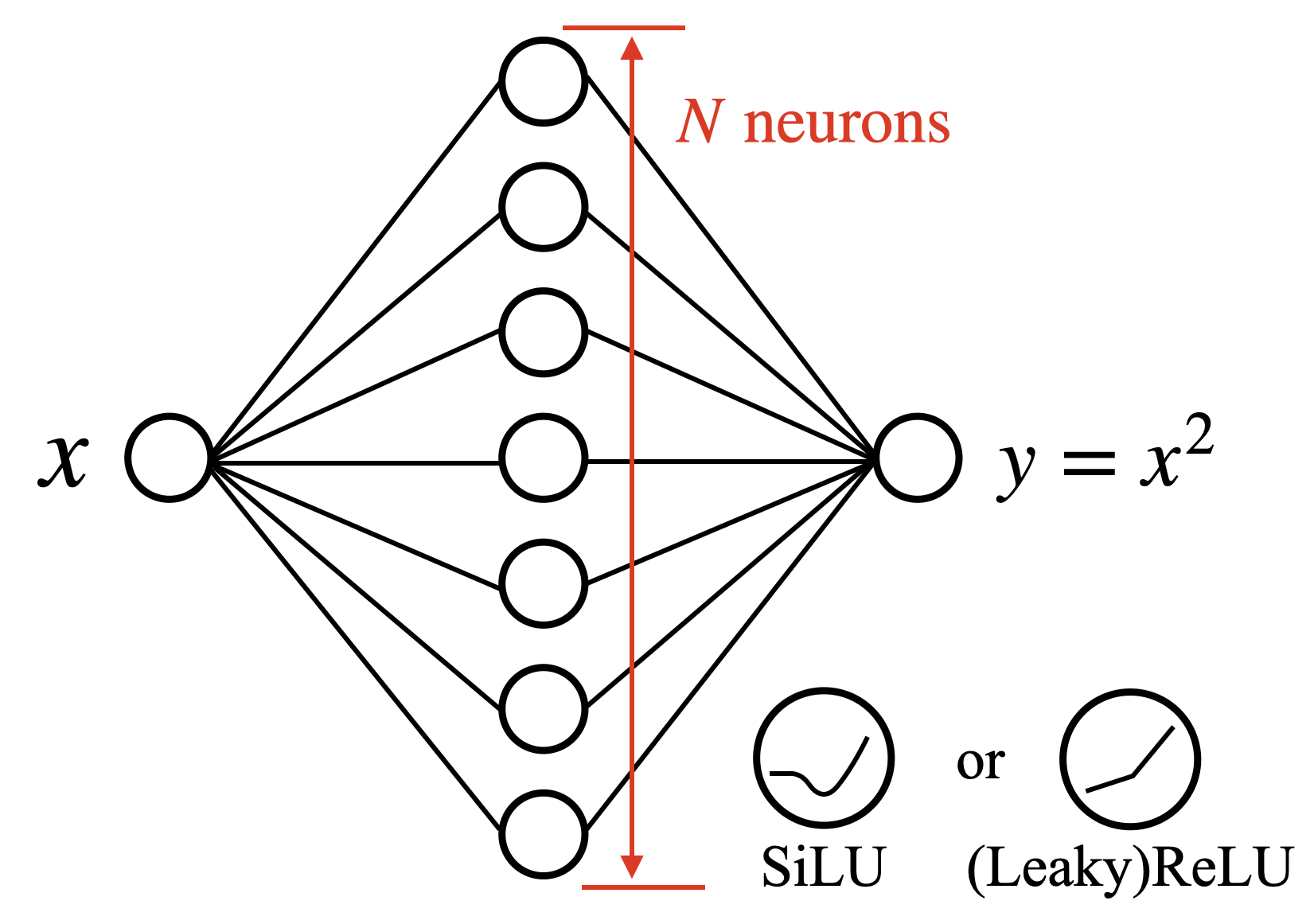}
    \caption{}
    \label{fig:NN}
    \end{subfigure}
    \begin{subfigure}[]{0.3\textwidth}
    \includegraphics[width=\linewidth, trim=0cm 0cm 0cm 0cm]{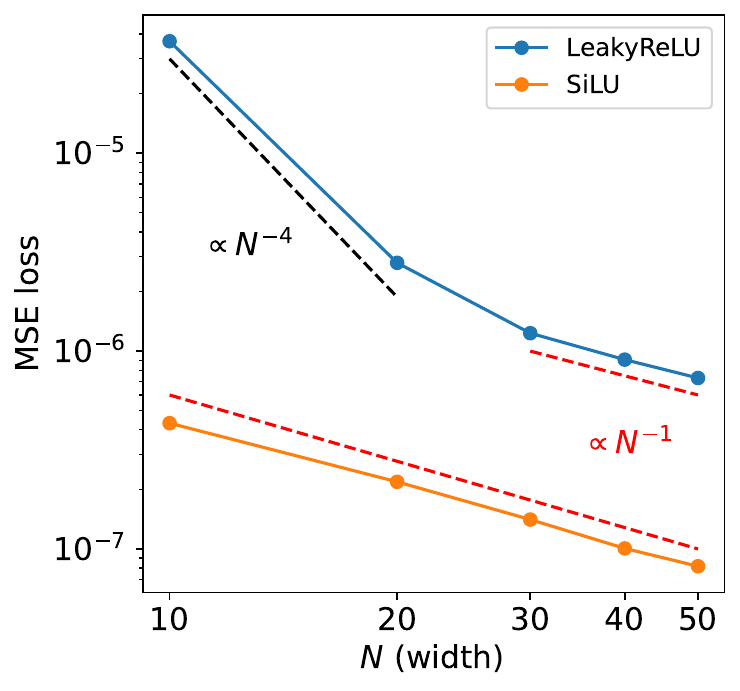}
    \caption{}
    \label{fig:scaling}
    \end{subfigure}
    \caption{(a) The $\ell\propto N^{-4/d}$ scaling law from Sharma and Kaplan~\cite{sharma2020neural} can be understood from approximating a $d$-dimensional function while data points lie uniformly inside a hypercube. (b) Our simple setup is training a two-layer SiLU or (Leaky)ReLU network (one hidden layer with width $N$) to fit the squared function $y=x^2$. (c) A surprising $N^{-1}$ scaling emerges for SiLU networks and at the tail of ReLU networks, while \cite{sharma2020neural}'s prediction $N^{-4}$ only appears at the early stage of ReLU.}
    \label{fig:competition}
    \vskip -0.4cm
\end{figure}

This paper aims to reveal yet another mechanism of neural scaling law from \textit{lottery ticket ensembling}. Our theory is motivated by empirical observations in an extremely simple setup, i.e., training two-layer (ReLU or SiLU) networks with the Adam optimizer to fit the squared function $y=x^2$. The approximation theory would predict that $\alpha=4$ for ReLU networks, but cannot predict $\alpha$ for SiLU networks. The quanta theory is also not applicable to these cases. Our empirical results are quite intriguing (shown in Figure~\ref{fig:scaling}): For ReLU networks, the loss curve decay as $N^{-4}$ at the beginning, but soon slows down to $N^{-1}$ as $N$ increases. For SiLU networks, the loss curve behaves as $N^{-1}$ consistently.

Now comes the question: what gives rise to this $N^{-1}$ scaling law? By reverse engineering single networks and doing statistical analysis on a population of networks, we attribute the new neural scaling law to \textit{lottery ticket ensembling}: a network with width $N$ contain $n$ "lottery tickets" ($n\appropto N$), ensembling of which can reduce the variance of outputs as $n^{-1}\appropto N^{-1}$. Lottery tickets refer to sub-networks which can, by their own, achieve good performance~\cite{frankle2018the}. The idea of ensembling is similar to bagging in machine learning where weak learners are aggregated to obtain strong learners, although in neural networks such ensembling strategy is not designed manually but rather emergent from training.

The paper is organized as follows: In Section \ref{sec:new_scaling_law}, we review the approximation theory and show our empirical results manifesting a scaling law $\ell\propto N^{-1}$ deviating from the approximation theory. To understand the new scaling law, in Section \ref{sec:lt} we reverse engineer neural networks to find out why. Hinted by the observation of symmetric neurons and peaks in the loss histogram, we find the existence of "lottery tickets". Then in Section \ref{sec:clt} we observe a central limit theorem for lottery tickets, which attributes the $N^{-1}$ scaling law to variance reduction as in central limit theorem. Finally in Section~\ref{sec:related-works} we discuss our theory's implications for large language models and statistical physics-like theories of learning.

\section{A New Scaling Law Not Explained by Approximation Theory}\label{sec:new_scaling_law}

In this section, we first review the prediction of NSL by Sharma \& Kaplan~\cite{sharma2020neural} using the approximation theory. Then we show one simple example which demonstrates a neural scaling law deviating from the approximation theory.

\subsection{The Old Tale: Approximating Functions On Data Manifold}

Sharma \& Kaplan~\cite{sharma2020neural} predicts that, when data is abundant, the MSE loss achieved by well-trained ReLU neural networks scales as $N^{-\alpha}$ with $\alpha=4/d$, where $d$ is the intrinsic dimensionality of data manifold. The basic idea is: A ReLU network with $N$ parameters is able to fit $O(N)$ data points accurately. If these  data points are uniformly placed on the grid of a $d$-dimensional hypercube (see Figure~\ref{fig:hypercube}), then there are $O(N^{1/d})$ points along each dimension, with lattice constant $h=O(N^{-1/d})$. ReLU networks are piecewise linear functions, so the leading error term (according to Taylor expansion) is second-order, i.e., $h^2$. Considering the squared function in the definition of MSE, we know the MSE loss scales as $h^4=O(N^{-4/d})$. 

\subsection{Experiments: Discovery of A New Scaling Law}
{\bf Setup} Let us consider a simple example. The network has only one hidden layer with $N$ neurons, with a scalar input $x\in [-2,2]$ and aiming to predict $y=x^2$ (Figure~\ref{fig:NN}, more examples in Appendix~\ref{app:examples}). The network is parametrized as
\begin{equation}
    f(x) = \sum_{i=1}^N v_i\sigma(w_ix+b_i) + c,
\end{equation}
where $\sigma$ is the activation function. When the activation function is ReLU, the approximation theory predicts that the MSE loss scales as $N^{-4}$ since $d=1$. When the activation function is SiLU ($\sigma(x)=x/(1+e^{-x})$, a smoothed version of ReLU), no existing theory is able to predict the scaling law. We are thus curious to run empirical experiments to see: (1) Is the $N^{-4}$ scaling law valid in our case? (2) what is the scaling law for SiLU networks?

\begin{figure}[tbp]
    \centering
    \includegraphics[width=1\linewidth]{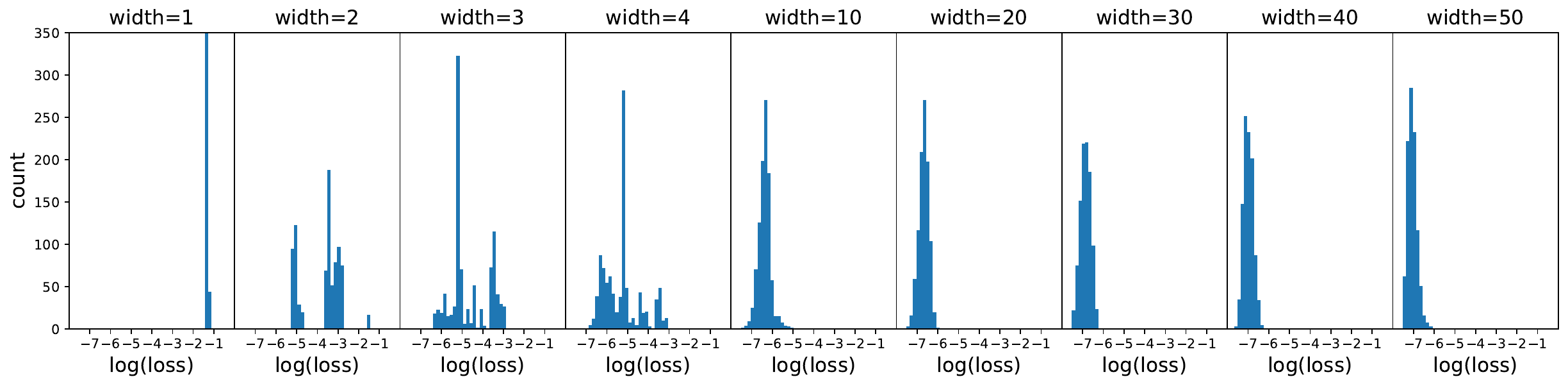}
    \caption{"Central limit theorem" of lottery tickets. For each width $N$, we train 1000 networks independently and plot their loss histograms. For small $N$, the distribution is multi-modal, i.e., shows more than one peaks; for large $N$, the distribution becomes more single-peaked.}
    \label{fig:loss_hist_many}
    \vskip -0.4cm
\end{figure}

{\bf Empirical Results} We train neural networks with various width $N=\{10,20,30,40,50\}$ for LeakyReLU and SiLU activations, respectively. We train neural networks with the Adam optimizer for 50000 steps; the initial learning rate is $10^{-2}$, reducing the learning rate by $\times 0.2$ every 10000 steps. Since neural networks can be sensitive to initializations, we train 1000 networks with different random seeds and use the median loss value (histograms are shown in Figure~\ref{fig:loss_hist_many}). The MSE losses versus width $N$ is shown in Figure~\ref{fig:scaling}. For ReLU, we find that the loss starts off as $N^{-4}$, agreeing with the approximation theory, but then quickly transits to a much slower decay $N^{-1}$. By contrast for SiLU, the MSE loss decays slowly but consistently as $N^{-1}$. The $N^{-1}$ scaling law is unexpected, suggesting a new mechanism not discovered before. In the next section, we aim to understand this new scaling law, by reverse engineering SiLU networks.

{\bf Remark: Optimization} Notice that the SiLU function $\sigma(x)=x/(1+e^{-x})$ expands to be a quadratic function around its minimum $\sigma(x)\approx A(x-x_*)^2+B$~\footnote{$A\approx 0.109, x_*\approx 1.278, B=-0.278$.}, so ideally one can carefully construct parameters of a $N=1$ network to make it approximate the squared function arbitrarily well, which requires the weights $w$ and $v$ to be $w\to 0$, $v\to+\infty$ but maintains $w^2v=1/A$ constant. In practice, optimization cannot find such extreme solution, otherwise loss would become perfectly zero even for $N=1$.

\section{Mechanistically Understanding Lottery Tickets}\label{sec:lt}

\begin{figure}[t]
    \centering
    \begin{subfigure}[]{0.3\textwidth}
    \includegraphics[width=\linewidth, trim=0cm 0cm 0cm 0cm]{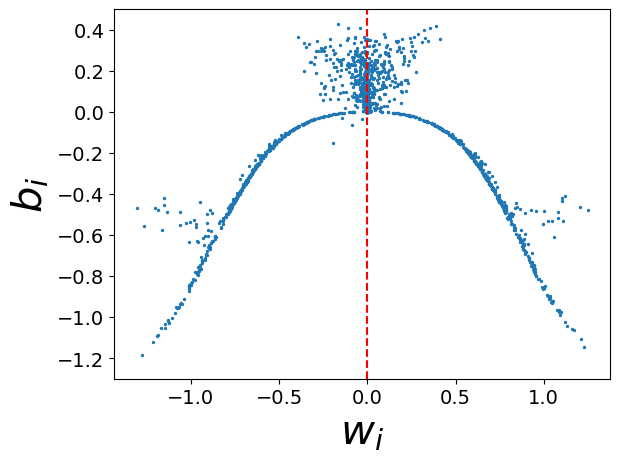}
    \caption{}
    \label{fig:wb_distribution}
    \end{subfigure}
    \begin{subfigure}[]{0.3\textwidth}
    \includegraphics[width=\linewidth, trim=0cm 0cm 0cm 0cm]{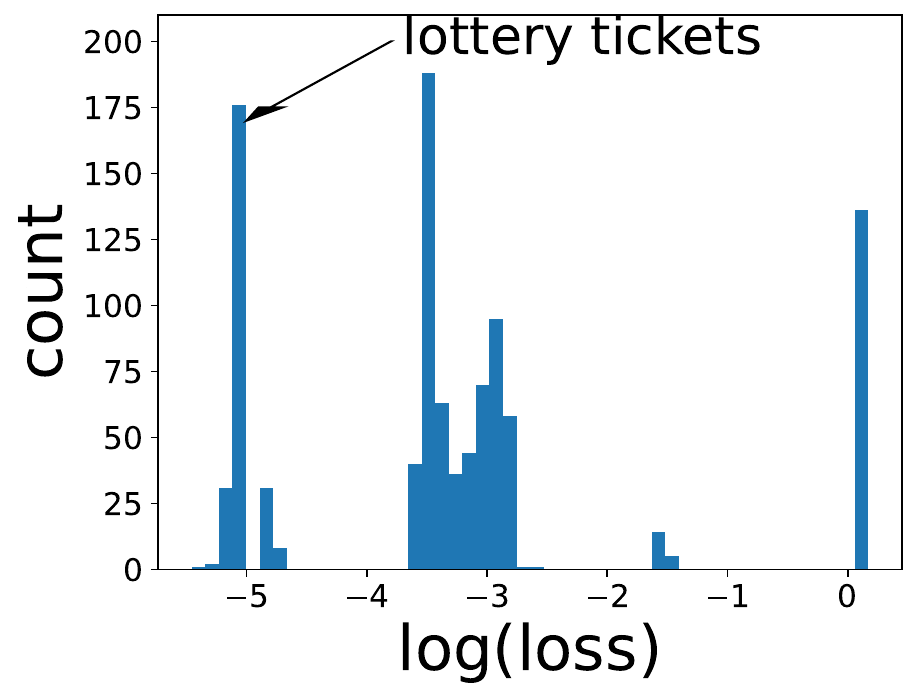}
    \caption{}
    \label{fig:loss_hist_2}
    \end{subfigure}
    \begin{subfigure}[]{0.35\textwidth}
    \includegraphics[width=\linewidth, trim=0cm 0cm 0cm 0cm]{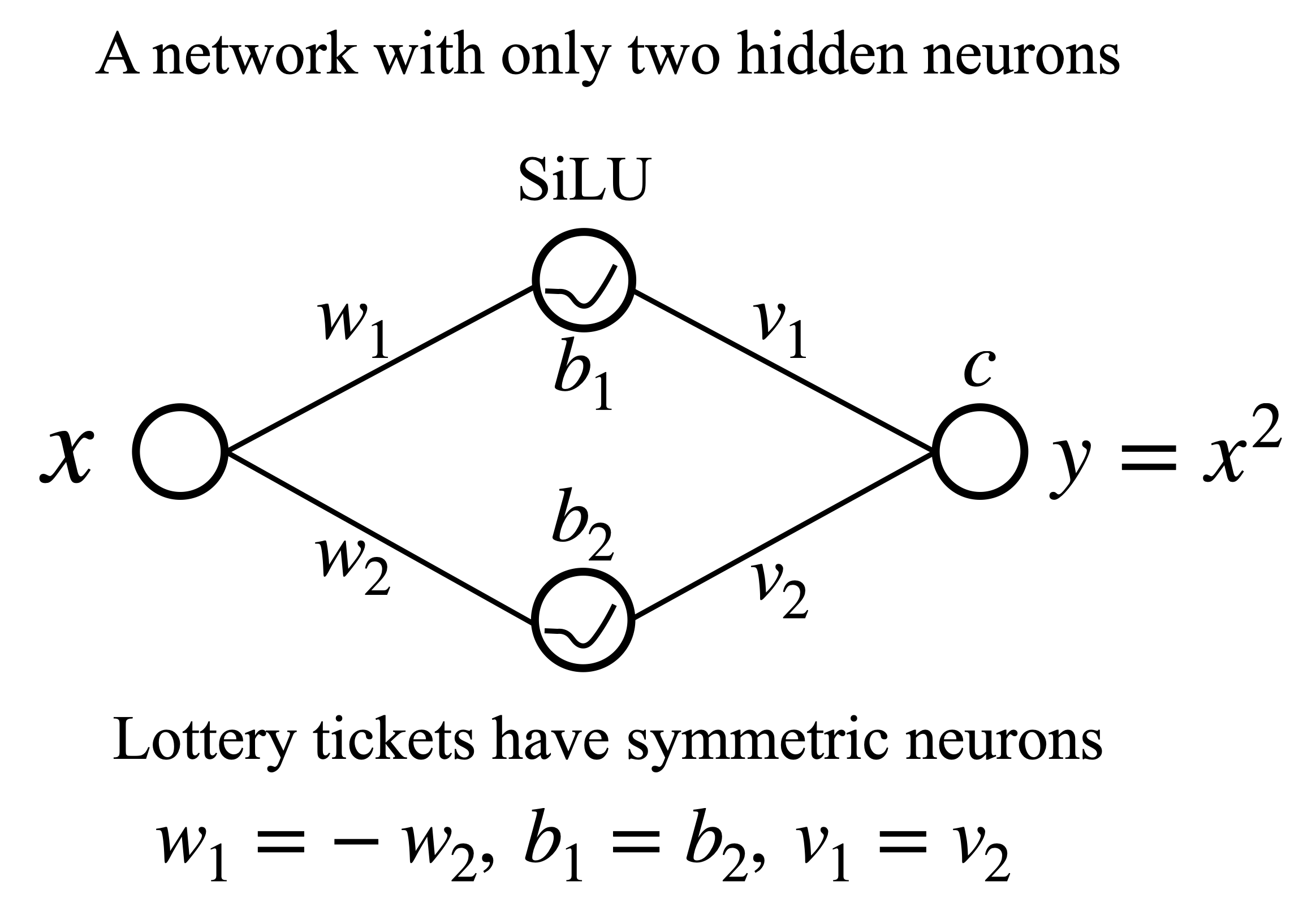}
    \caption{}
\label{fig:illust_lt}
    \end{subfigure}
    \caption{Evidence of lottery tickets. (a) For an extremely wide network $N$=10000, the distribution of weights and biases in the first layer display an intriguing symmetry, i.e., there exist symmetric neurons $(w,b)$ and $(-w,b)$. (b) We train a thousand $N=2$ networks independently and show the histogram of their losses. The histogram display a few peaks, suggesting existence of a few different  local minima or "algorithms". We call the peak with lowest loss "lottery tickets". (c) We find lottery tickets to have symmetric neurons, which guarantee that the network represents an even function.}
    \label{fig:lt-evidence}
    \vskip -0.4cm
\end{figure}

{\bf Existence of symmetric neurons in wide networks} To understand what is happening inside a wide network, we train an extremely wide network with $N=10000$ SiLU neurons.  We plot $N$ neurons $(w_i,b_i)$ (weights and biases for the first layer) for $i=1,\cdots,N$ in Figure~\ref{fig:wb_distribution}. There are two interesting patterns: (1) most neurons are concentrated around a bell-shaped curve, indicating an attractor manifold; (2) the distribution is symmetric with respect to $w$: if there is a neuron at $(w,b)$, then there is a "symmetric" neuron around $(-w,b)$. The existence of the attractor manifold may not be too surprising, since low-loss solutions should live in a subspace smaller than the entire parameter space. The existence of symmetric neurons, however, is somewhat unexpected and requires more detailed study, as we carry out below.

{\bf Two-neuron networks} We conjecture that, if symmetric neurons are universal enough, we should be able to observe them even in a network with 2 hidden neurons. Hence we train SiLU networks with only 2 hidden neurons to fit the squared function (see Figure~\ref{fig:illust_lt}). Since such narrow networks are strongly affected by initializations, we train 1000 networks with different random seeds. We show the histogram of MSE losses of these trained networks in Figure~\ref{fig:loss_hist_2}, finding that it contains many peaks, suggesting the existence of bad local minima in loss landscapes. In particular, we are interested in the peak with the lowest loss, which we call "lottery tickets", elaborated below. We also attempt to understand other peaks in parameter space and in algorithmic space in Appendix~\ref{app:mi}.

A randomly chosen lottery ticket has parameters (illustrated in Figure~\ref{fig:illust_lt}):
\begin{equation}
\begin{aligned}
    &(w_1, w_2, b_1, b_2, v_1, v_2, c) 
    \\ = &(-0.83010483,  0.83010304, -1.16842330, -1.16842365,  5.68583536,
         5.68586636,  3.1539197)
\end{aligned}
\end{equation}
where we observe $w_1\approx -w_2$, $b_1\approx b_2$, $v_1\approx v_2$, which correspond to symmetric neurons. The benefits of symmetric neurons can be understood from Taylor expansion. Setting $w_1=-w_2=w$, $b_1=b_2=b$, $v_1=v_2=v$, the network represents an even function
\begin{equation}
    f(x) = v\sigma(wx+b) + v\sigma(-wx+b) + c,\quad f(x) = f(-x),
\end{equation}
and Taylor expands as 
\begin{equation}
    f(x) = (2\sigma(b)u+c) + u\sigma''(b)w^2x^2 +  \frac{1}{12}u\sigma''''(b)w^4x^4 + ...,
\end{equation}
so a neural network can adjust its parameters to make sure $2\sigma(b)u+c=0$, $u\sigma''(b)w^2=1$, leaving the quartic term as the leading error term. The take-away is: utilizing symmetric neurons is very effective at approximating the squared function, however, it relies on luck for networks to find such strategy (justified the terminology "lottery tickets").

\section{A Central Limit Theorem of Lottery Tickets}\label{sec:clt}

In the previous section, we show the loss histogram of width $N=2$ networks contains many peaks, and define the peak with the lowest loss as "lottery tickets". How would the story change for wider networks (i.e., larger $N$)? 

{\bf Loss histograms} For each width $N=1,2,3,4,10,20,30,40,50$, we train 1000 networks independently (with different random seeds) and plot their loss histograms in Figure~\ref{fig:loss_hist_many}. For narrow networks ($N=1,2,3,4$), the loss histogram contains many peaks, suggesting the existence of many bad local minima, but on the other hand, suggesting the existence of "lottery tickets" which win over other local minima. For wider networks ($N=10,20,30,40,50$), the loss histograms are single-peaked. This reminds us of the central limit theorem of random variables: the average of arbitrary weird-shaped distribution would converge to a Gaussian distribution as more and more random variables are averaged. This suggests that maybe there are many lottery tickets inside a wide network and then ensembled. Below we propose a theory for it.

{\bf Theory: Ensembling Lottery Tickets Leads to $N^{-1}$ Scaling}
For a wide nertwork, there exist $n>1$ lottery tickets. The $i^{\rm th}$ lottery ticket represents a function $f_i(x)\approx f(x)$ whose error term $e_i(x)\equiv f_i(x)-f(x)$. We define function norm as $|f|^2=\int_{-\infty}^\infty f^2(x)p(x)dx$~\footnote{In our toy model, $x$ is drawn uniformly from $[-1,1]$, so $p(x)=1$ for $x\in[-1,1]$, and $p(x)=0$ otherwise. $|f|^2=\int_{-1}^1 f^2(x)dx$.}, where $p(x)\geq 0$ is the distribution of $x$. Typically $|e_i(x)|^2\ll|f|^2$. The network can utilize the last linear layer to ensemble these lottery tickets such that
\begin{equation}
    f_{E}(x) = \sum_{i=1}^n a_i f_i(x) = (\sum_{i=1}^n a_i)f(x) + \sum_{i=1}^n a_ie_i(x).
\end{equation}
 We want to minimize:
\begin{equation}
    |f_E(x)-f(x)|^2 = \left|(\sum_{i=1}^n a_i-1)f(x) + \sum_{i=1}^n a_ie_i(x)\right|^2.
\end{equation}
If we assume $f(x)$ to be orthogonal to $e_i(x)$, and $e_i(x)$ is orthogonal to $e_j(x)$ for $j\neq i$ (in general we don't need orthogonality conditions~\footnote{If $\langle e_i, e_j\rangle\neq 0$, i.e., they are not orthogonal, we can always redefine $e'_j\equiv e_j - \frac{\langle e_j,e_i\rangle}{\langle e_i, e_i\rangle}e_i$ such that $\langle e_i, e'_j\rangle=0.$ In summary, we can always make lottery tickets to be orthogonal in the proper bases. However, $k$ lottery tickets may actually have only $n < k$ bases. In fact, lottery tickets are highly correlated (as we show in Appendix~\ref{app:disentangle}), so here $n$ should be understood as the number of independent/orthogonal lottery tickets.}), then the above equation simplifies to
\begin{equation}
    |f_E(x)-f(x)|^2=(\sum_{i=1}^n a_i-1)^2 |f|^2 + \sum_{i=1}^n a_i^2|e_i|^2
\end{equation}
Since $|f|^2\gg |e_i|^2$, we want $\sum_{i=1}^N a_i=1$ such that the coefficient before $|f|^2$ becomes zero. The second term can be lower bounded by (using Cauchy-Schwarz inequality):
\begin{equation}
    (\sum_{i=1}^n a_i^2|e_i|^2)(\sum_{i=1}^n \frac{1}{|e_i|^2}) \geq (\sum_{i=1}^n a_i)^2 = 1
\end{equation}
The equality holds when $a_i|e_i|^2 = C$ for all $i$, meaning that for a better/worse lottery ticket (small/large $|e_i|^2$), a larger/smaller weight $a_i$ is applied. Finally we arrive that
\begin{equation}
    |e_{E}|^2\equiv |f_E(x)-f(x)|^2 \geq (\sum_{i=1}^n \frac{1}{|e_i|^2})^{-1}
\end{equation}
If we further assume $n$ lottery tickets to be of equal quality, i.e., $|e_i|^2=|e|^2$, then $|e_E|^2=|e|^2/n$, meaning that the MSE loss decays as $n^{-1}$. A width $N$ network on average has $n\propto N$ lottery tickets, so the $n^{-1}$ scaling can further translate to $N^{-1}$.

\begin{figure}[tbp]
    \centering
    \includegraphics[width=0.95\linewidth]{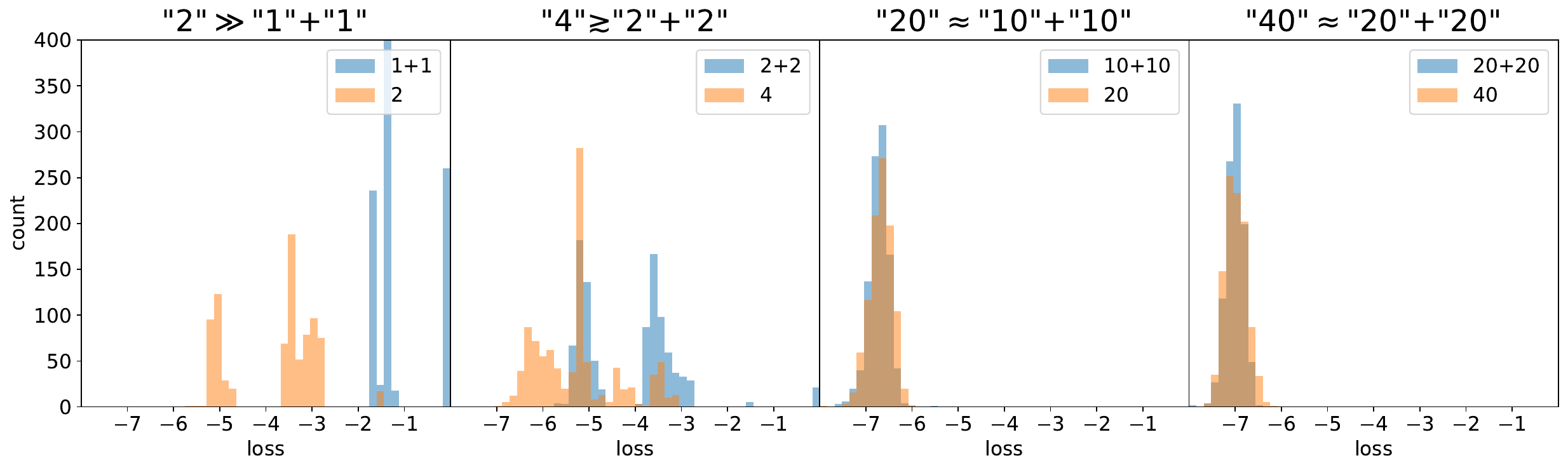}
    \caption{Do the benefits of large widths come from  plain ensembling or more complicated synergy of smaller subparts? In each plot "$N$" means a network with width $N$, "$N/2$"+"$N/2$" means two networks with width $N/2$ are ensembled. If these two loss histograms are different (e.g., $N=2,4$), this means more complicated synergy is in place beyond ensembling. If the two loss histograms are similar (e.g., $N=20,40$), this means the role of synergy is vanishing, and the benefit of larger widths solely comes from ensembling.}
    \label{fig:lt_id}
    \vskip -0.4cm
\end{figure}

{\bf The benefits of large widths from ensembling or synergy} Ensembling is only one benefit of wider models. In some sense, ensembling is trivial because you can train two half-sized models independently and then ensemble them. Are there benefits of large widths beyond ensembling, i.e., two sub-networks might synergize in a smart way that smaller networks cannot simulate? To see the benefits of synergy, we need to first subtract the effect of ensembling. The loss of networks $N/2$ and $N$, obeys distributions $L_{N/2}\sim p_{N/2}(\ell)$ and $L_N\sim p_N(\ell)$, respectively. Ensembling two $N/2$ networks, according to our theory above, would give loss $\tilde{L}_N=(1/L_{N/2}+1/L_{N/2})^{-1}$, obeying the distribution $\tilde{p}_N(\ell)$. We then compare the difference between $\tilde{p}_N(\ell)$ and $p_N(\ell)$; larger (smaller) difference means more (less) synergy between two $N/2$ networks. In practice, we obtain the histogram of $\tilde{p}_N(l)$ by randomly drawing two losses from the histogram of $\tilde{p}_{N/2}(l)$ and compute the harmonic mean (divided by 2). 

Our empirical results are shown in Figure~\ref{fig:lt_id}: narrower networks have stronger synergy, while wider networks are almost ensembling with nearly no synergy. For example, an $N=2$ network is much better than ensembling of two $N/2=1$ networks ($2\gg 1+1$); an $N=4$ network is (slightly) better than the ensembling of two $N/2=2$ networks ($4 > 2+2$); an $N=20$ or $N=40$ network does as well as ensembling of two half-width networks ($20\approx 10+10, 40\approx 20+20$). Although in Section \ref{sec:lt}, we only analyzed $N=2$ lottery tickets, $N=4$ lottery tickets are even better, i.e., one cannot construct an $N=4$ lottery ticket from simply ensembling two $N=2$ lottery tickets. This may imply that wider networks may not only provide {\bf more} lottery tickets, but also provide {\bf better} lottery tickets when synergy is present.

{\bf Implications for large language models} Lottery ticket ensembling would expect the loss to be $\ell\propto w^{-1}$ where $w$ is the width of the network. For large language models, people use $N$ to represent the number of parameters, which is roughly $({\rm width})^2\times {\rm depth}$. Since the depth-width ratio~\cite{levine2020depth} is usually fixed when scaling language models, we have $w\propto N^{1/3}$ ($N$ is the number of parameters) hence $\ell\propto N^{-1/3}$. Interestingly, Hoffmann et al.~\cite{hoffmann2022training} reports that~\footnote{$D$ is the dataset size. In our case $D\to\infty$ so we can ignore the loss dependence on $D$.}:
\begin{equation}
    \ell(N,D) = E + \frac{A}{N^{0.34}} + \frac{B}{D^{0.28}},
\end{equation}
whose model size scaling exponent $0.34$ is quite close to our prediction $1/3$. We would like to investigate in the future whether this really supports our theory or it is a pure coincidence.

\section{Related works and Discussions}\label{sec:related-works}

{\bf Neural Scaling laws (NSL)} has been widely observed in deep learning ~\cite{hestness2017deep, rosenfeld2019constructive, kaplan2020scaling, henighan2020scaling, gordon2021data, zhai2022scaling, hoffmann2022training,michaud2023quantization}. Theories are proposed to explain NSL, including approximation theory based on intrinsic input dimension~\cite{sharma2020neural} or maximum arity~\cite{michaud2023precision}, quanta theory based on subtask decomposition~\cite{michaud2023quantization}. Our work prosposes another NSL mechanism namely lottery ticket mechanism. It is still unclear how to disentangle these mechanisms, and whether there is a dominant mechanism in deep networks. 

{\bf Emergence} refers to the phenomenon where a model has abrupt improvements in performance when its size is scaled up~\cite{wei2022emergent}, although this can depend on specific metrics~\cite{schaeffer2023emergent}. Our discussion on ensembling versus synergy may further provide a useful tool to categorize types of emergence: trivial (pure ensembling) or non-trivial (with synergy).  

{\bf Ensembling} is not new in machine learning. However, ensembling methods are usually manually designed~\cite{lobacheva2020power}. This work shows that ensembling can be emergent from network training. A very related concept is redundancy, which has been shown both for vision tasks~\cite{doimo2021redundant}, language tasks~\cite{mcgrath2023hydra}, and even simple math datasets~\cite{zhong2023clock,liu2023seeing}.

{\bf Mean field theory} suggests that infinitely wide neural networks approach kernel machines, and finite networks with finite width $N$ deviates from the limiting kernel by order $1/N$~\cite{halverson2021neural,roberts2022principles,rotskoff2018parameters}, which agrees with our lottery ticket central limit theorem. This is a pleasant agreement: our analysis starts from narrow networks (lottery tickets) and extends to wide networks, while mean field theory starts from infinitely wide networks and extends to  (finitely) wide networks. Unifying two theories would provide clearer mental pictures, left for future work.

{\bf Limitations} The derivation of the $N^{-1}$ scaling requires the  assumption that lottery tickets are unbiased. If all lottery tickets share a non-zero bias, then the loss would plateau to a non-zero value due to the bias term. We indeed observe this phenomenon for some cases in Appendix~\ref{app:examples}. The analysis in this paper is mainly based on a toy example; how general our claims are for other tasks and architectures need more study in the future.

\section*{Acknowledgement} 
We would like to thank Eric Michaud, Isaac Liao and Zechen Zhang for helpful discussions. ZL and MT are supported by IAIFI through NSF grant PHY-2019786, the Foundational Questions Institute and the Rothberg Family Fund for Cognitive Science. 


\bibliography{lt-nsl.bib}

\newpage
\appendix

{\huge Appendix}

\section{More examples}\label{app:examples}
Three key observations from the main paper for the $y=x^2$ example are: (1) The existence of lottery tickets for narrow networks (the loss histogram contains many peaks); (2) the loss histogram for wider networks has only one peak (central limit theorem); (3) the loss decays as $N^{-1}$ for wide networks ($N$ is the model width).

In this section, we would like to test these three claims on a few more examples: unary functions and multi-digit multiplication. Both (1) and (2) holds generally true for almost all examples, but (3) sometimes break down.

\begin{figure}[htbp]
    \centering
    \includegraphics[width=1\linewidth]{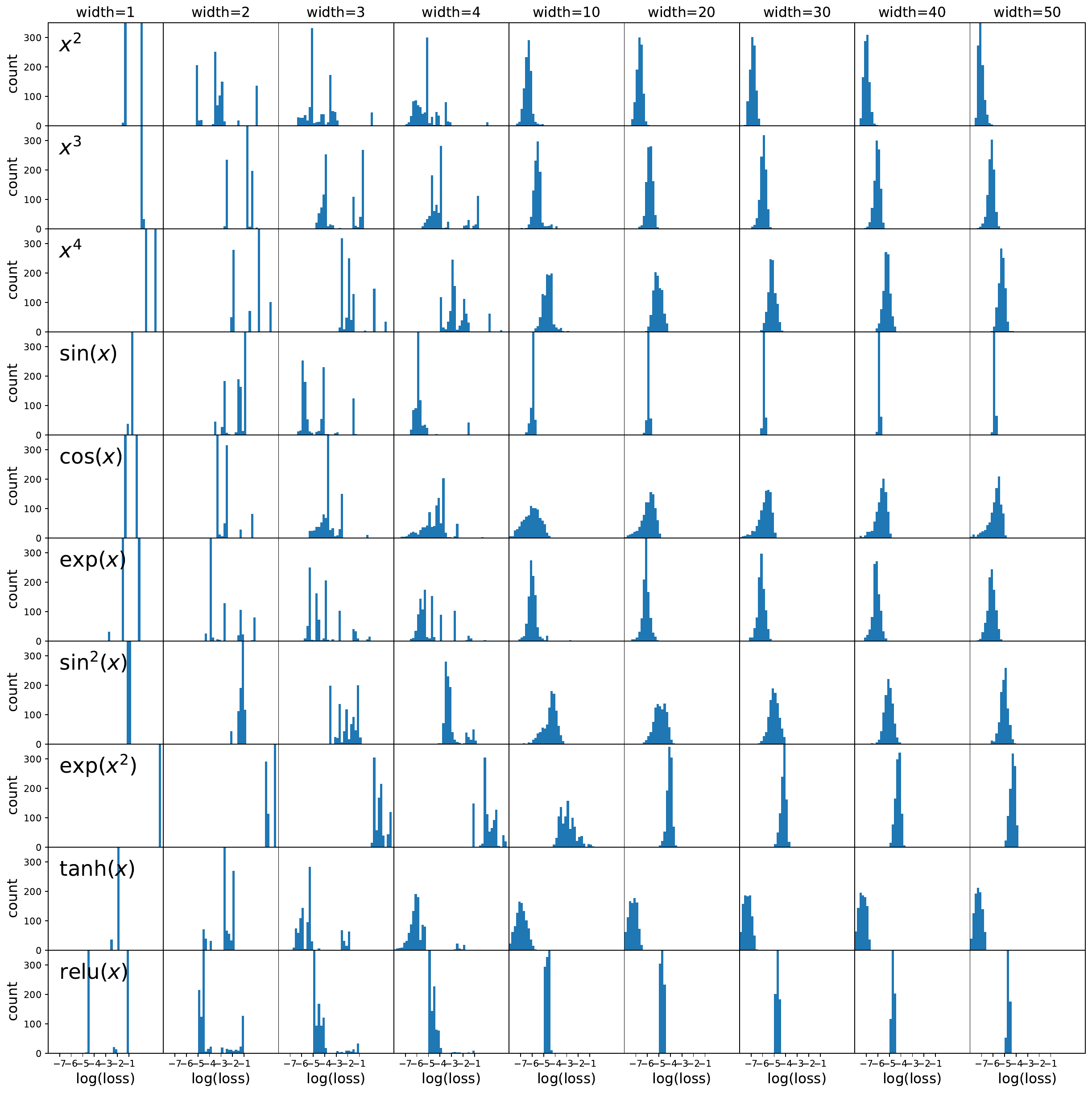}
    \caption{NN loss histograms for unary functions.}
    \label{fig:unary_hist}
\end{figure}

\begin{figure}[htbp]
    \centering
    \includegraphics[width=0.5\linewidth]{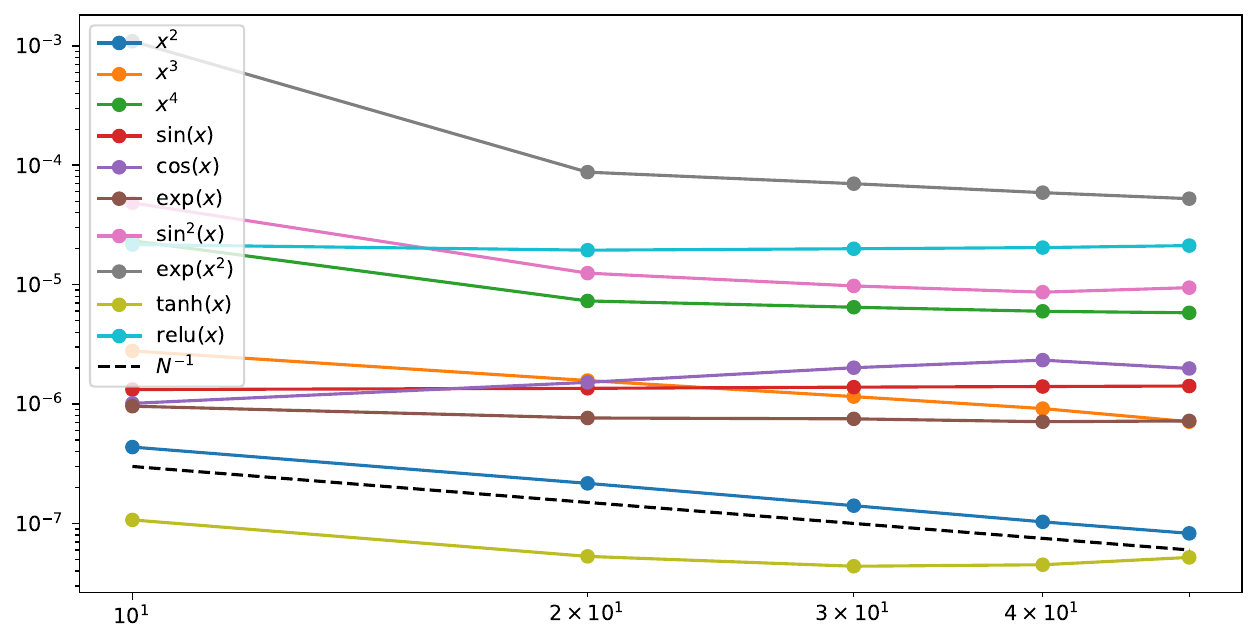}
    \caption{Neural scaling laws for unary functions.}
    \label{fig:unary_nsl}
\end{figure}

{\bf Unary functions} The experimental setups are exactly the same as in the main paper, with the only changes to target function $f(x)=x^3, x^4, {\rm sin}(x), {\rm exp}(x), {\rm sin}^2(x), {\rm exp}(x^2), {\rm relu}(x), {\rm tanh}(x)$. For each task, we run 1000 networks (with different random seeds, i.e., different initializations) and plot the histogram of their final losses in Figure~\ref{fig:unary_hist}. For narrow networks, the loss histogram always shows multiple peaks, implying the existence of lottery tickets (networks with lowest losses). As width grows larger, the loss histogram becomes single-peaked. We also have a few interesting observations (which we haven't been able to explain yet): (1) some examples are more asymmetric than others, e.g, ${\rm cos}(x)$ is very heavy-tailed towards the left. (2) for some examples the losse are more concentrated than other, e.g., ${\rm sin}(x)$ is extremly concentrated looking like a delta function. We compute the median value of 1000 losses for each width and plot the median loss as a function of width $N$ in Figure~\ref{fig:unary_nsl}. Some examples obey the $N^{-1}$ law quite well: $x^2$, $x^3$, ${\rm tanh}(x)$ (early stage); for some examples the losses are still decaying but seem to plateau, e.g., $x^4$, ${\rm sin}^2x$, ${\rm exp}(x^2)$; other examples already plateaued or even slightly increase. The two later scenarios are possibly due to the bias term of lottery tickets, or optimization issues for too wide networks.

\begin{figure}[htbp]
    \centering
    \includegraphics[width=1\linewidth]{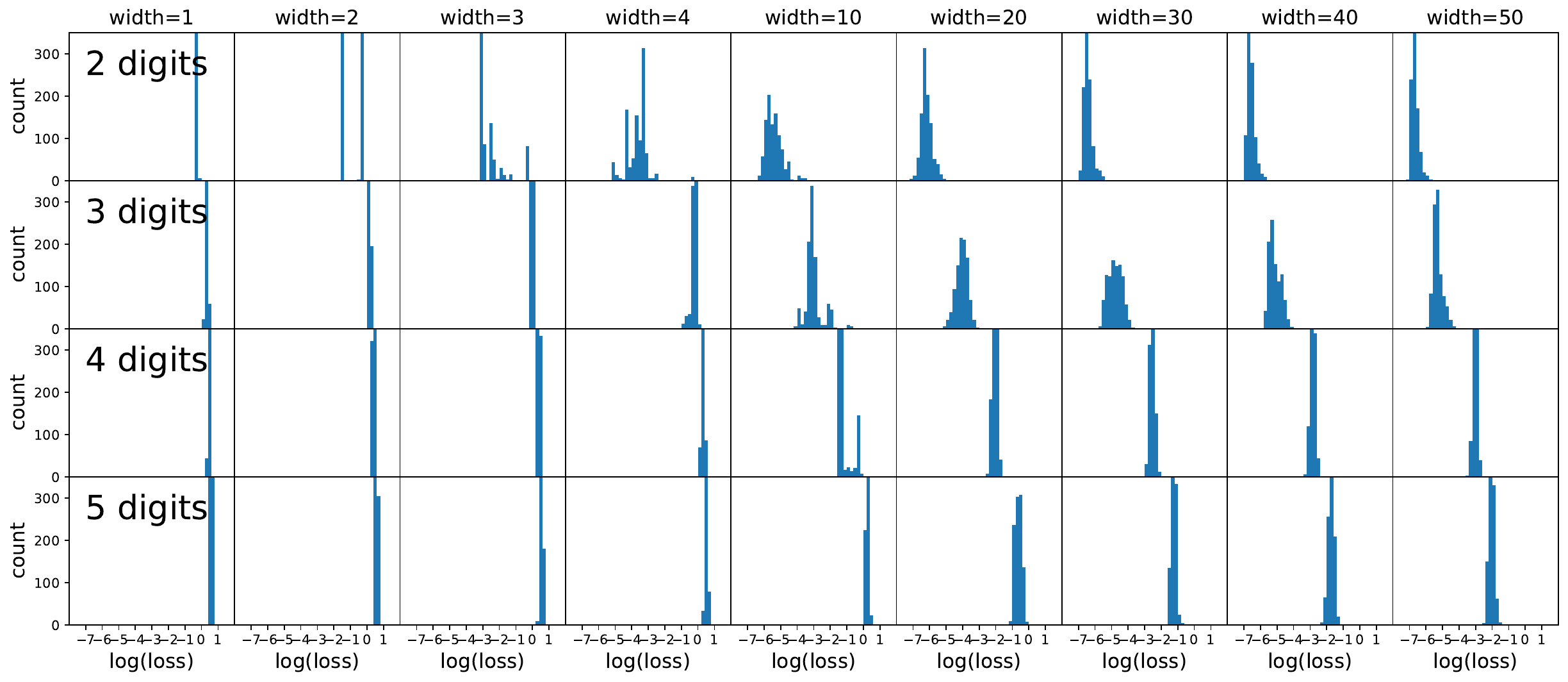}
    \caption{NN loss histograms for multiple-digit multiplication.}
    \label{fig:mult_hist}
\end{figure}

\begin{figure}[htbp]
    \centering
    \includegraphics[width=0.5\linewidth]{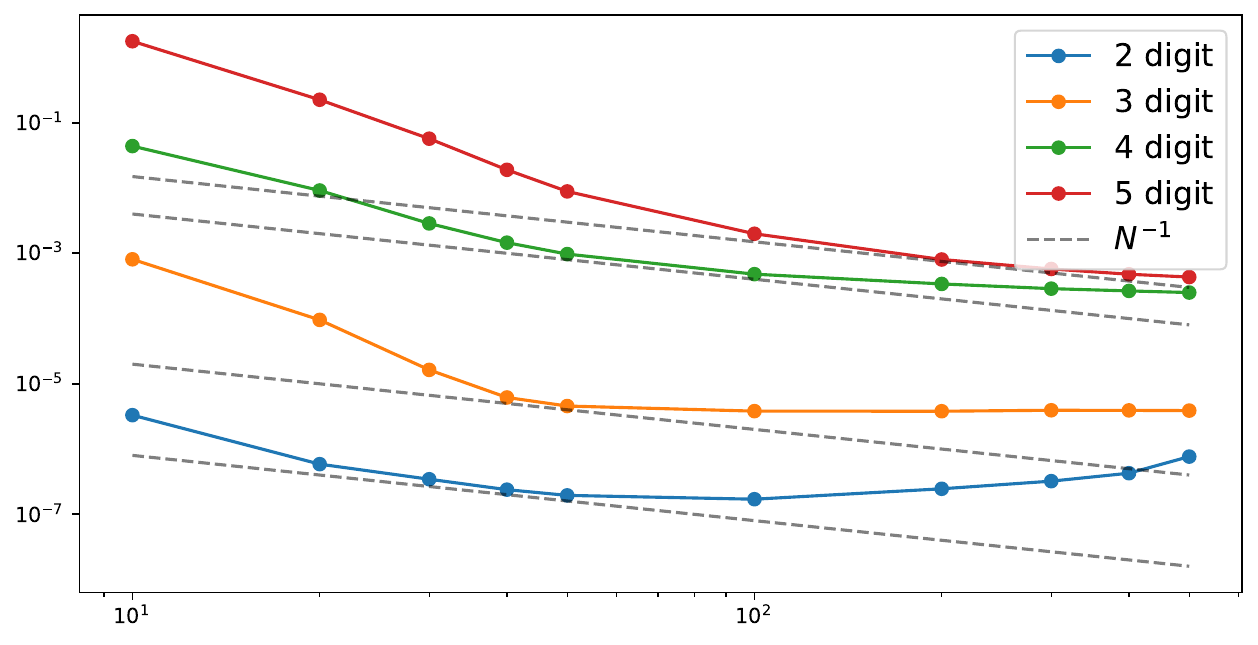}
    \caption{Neural scaling laws for multiple-digit multiplication.}
    \label{fig:mult_nsl}
\end{figure}

{\bf Multi-digit multiplication} We extend our input from one dimension to multiple dimensions. We do this multi-digit multiplication tasks with 2, 3, 4 or 5 digits, i.e., $f=x_1x_2, x_1x_2x_3, x_1x_2x_3, x_1x_2x_3x_4, x_1x_2x_3x_4x_5$. For each task, we run 1000 networks with different random seeds and plot their loss histograms in Figure~\ref{fig:mult_hist}. For narrow networks, the 2-digit case shows multiple peaks early on (e.g., width=2,3,4), but the 3-digit or 4-digit case show multiple peaks only for width=10, and 5-digit case do not show any multiple peak behavior in the ranges shown. We conjecture that this is because networks with only one hidden layer are very inefficient at multiplying numbers, especially for more numbers to be multiplied, and consequently, there is no (clear) lottery ticket for the ranges shown (up to 50). Regarding neural scaling laws, we plot the median loss as a function of $N$ in Figure~\ref{fig:mult_nsl}. For all cases (2,3,4,5 digits), at least some range of the loss curves agree with the $N^{-1}$ scaling law, where the lottery ticket ensembling mechanism is dominant. Before the range the loss drops faster because better lottery tickets are formed (synergy is in place); after the range the loss drops more slowly, plateaus or even increases due to biases of lottery tickets and optimization issues.

\section{Algorithm discovery by parameter space clustering}\label{app:mi}

In Figure~\ref{fig:loss_hist_2}, we showed that the loss histogram contains multiple peaks. If we only care about peaks below loss $10^{-2}$, then there are 4 peaks. What do these peaks mean in terms of model parameter space and algorithmic space? We will investigate these questions below. 

\begin{figure}[h]
    \centering
    \includegraphics[width=0.7\linewidth]{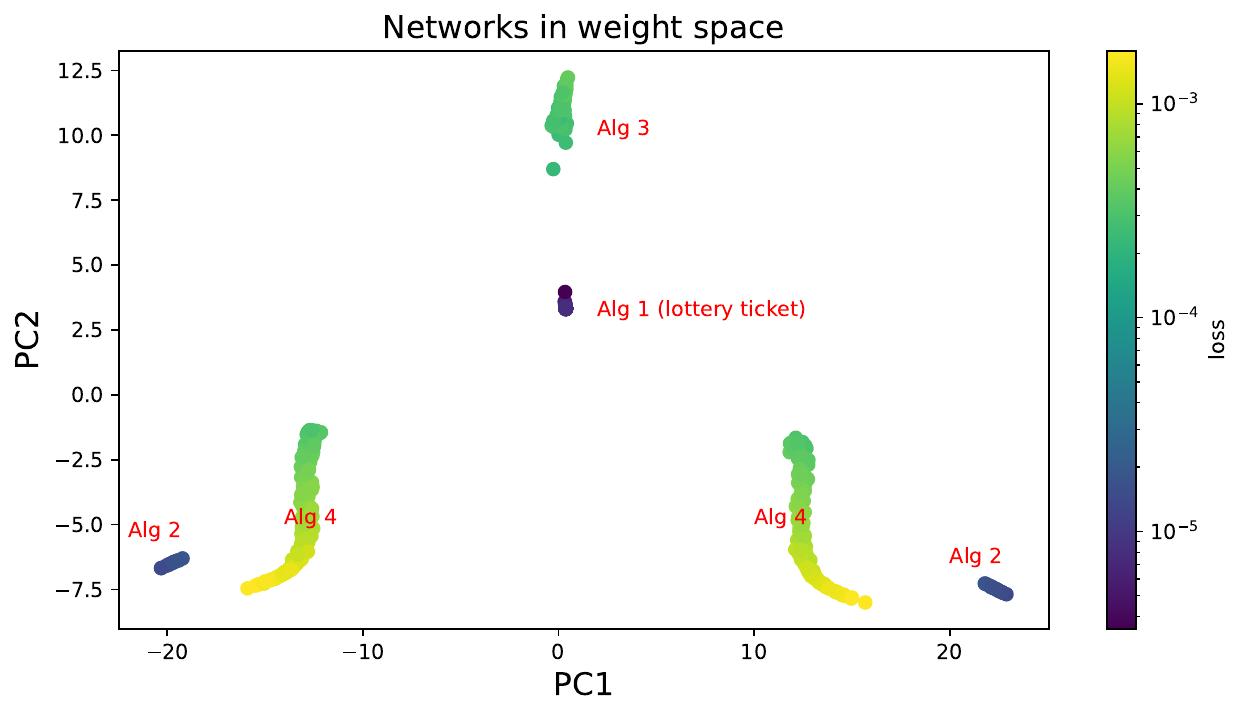}
    \caption{We train 1000 networks with only 2 hidden neurons to fit $f(x)=x^2$. Each network has 7 parameters hence is a point in $\mathbb{R}^7$. Applying principal component analysis (PCA) to the 1000 points, showing the first two PCs, and clusters are reavealed. Each culster corresponds to a distinct "algorithm".}
    \label{fig:mi_weight}
\end{figure}

{\bf Parameter space} Remember that we trained $1000$ width 2 neural networks. Each network contains 7 parameters, so it can be viewed as a point in the parameter space $\mathbb{R}^7$. We apply principal component analysis (PCA) to the 1000 networks in the parameter space, and show them on the first two principal components in Figure~\ref{fig:mi_weight}. Clearly there are 6 clusters, which are symmetric with respect to the first principal component. We verify that the symmetry comes form the permutation symmetry of two neurons (swapping two neurons does not alter the function represented by the network). Modulo the permutation symmetry, there are 4 distinct clusters. On the low level, the fact of more than one clusters implies the existence of local minima; on the high level, presumbly each cluster corresponds to an algorithm, which we elaborate below.

{\bf Algorithmic space} One thing we already noted is that: Algorithm 1 and 3 lie on the symmetric axis, meaning they are symmetric algorithms (two neurons play symmetric roles), while algorithm 2 and 4 are asymmetric. 

\begin{figure}[h]
    \centering
    \begin{subfigure}[]{0.23\textwidth}
\includegraphics[width=\linewidth, trim=0cm 0cm 0cm 0cm]{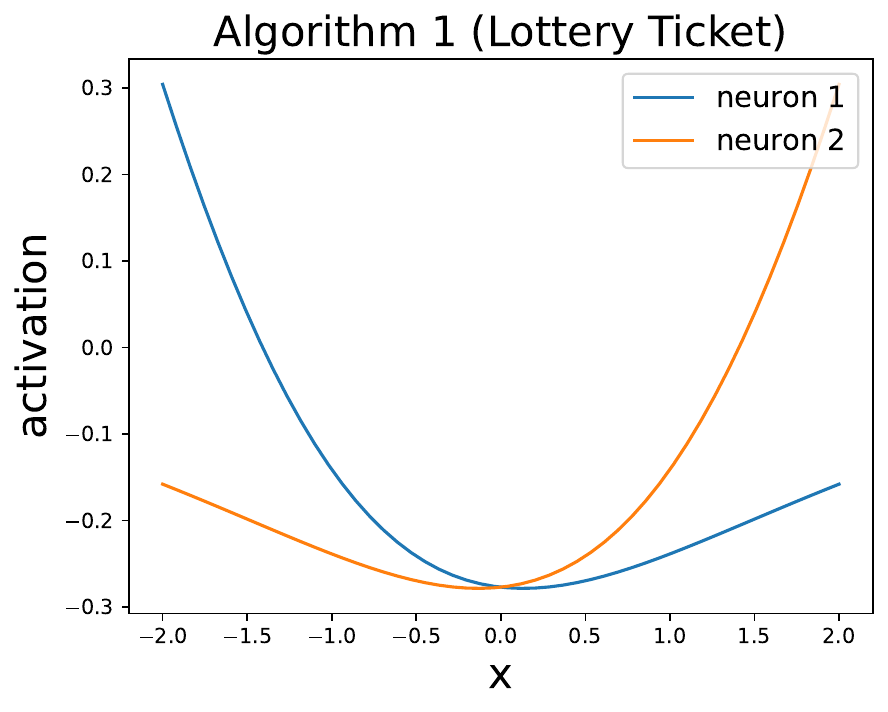}
    \caption{}
    \end{subfigure}
    \begin{subfigure}[]{0.23\textwidth}
    \includegraphics[width=\linewidth, trim=0cm 0cm 0cm 0cm]{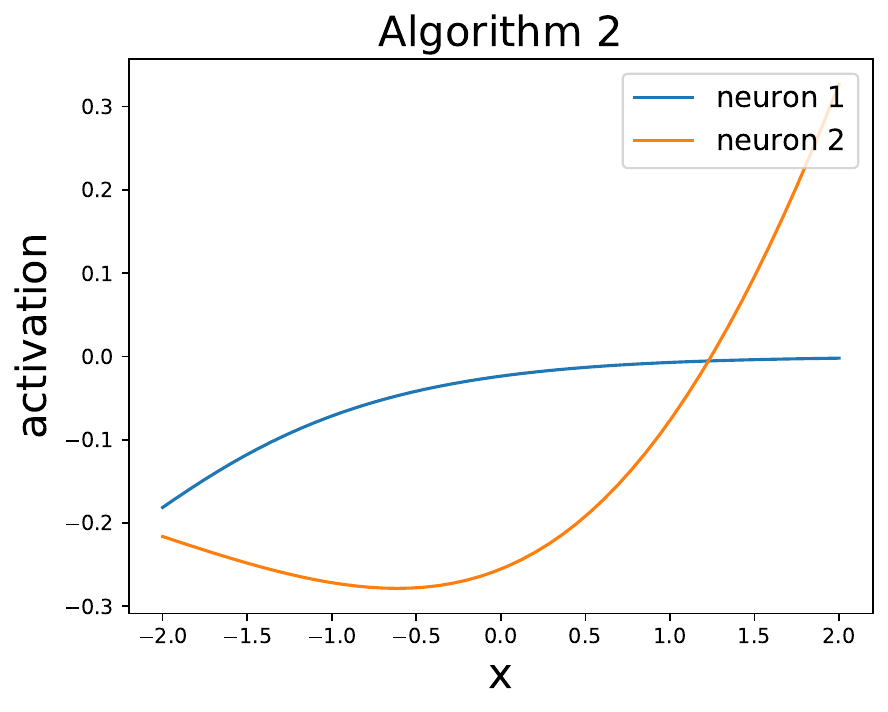}
    \caption{}
    \end{subfigure}
    \begin{subfigure}[]{0.23\textwidth}
    \includegraphics[width=\linewidth, trim=0cm 0cm 0cm 0cm]{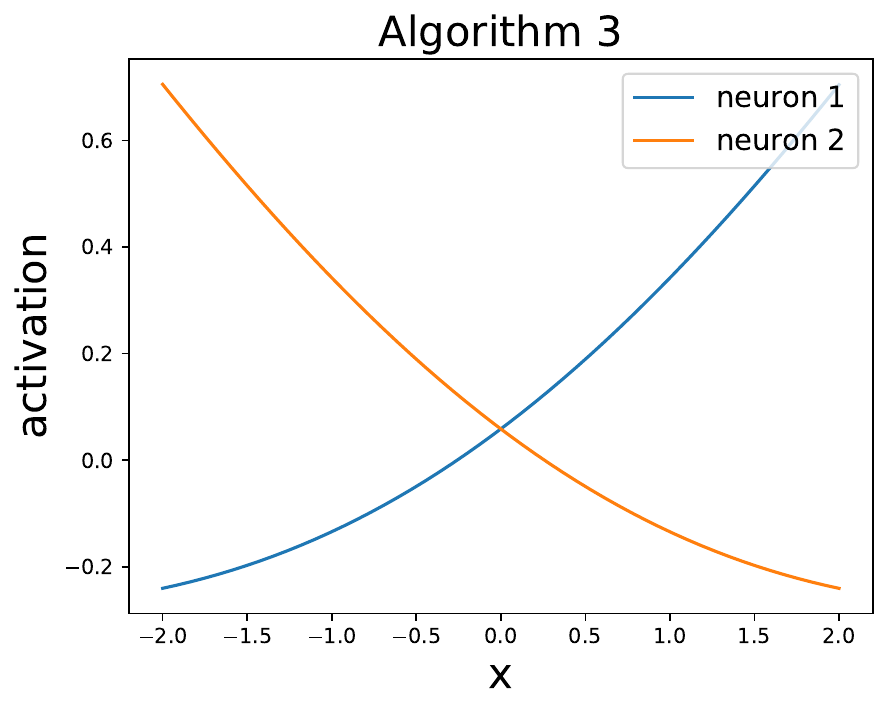}
    \caption{}
    \end{subfigure}
    \begin{subfigure}[]{0.23\textwidth}
    \includegraphics[width=\linewidth, trim=0cm 0cm 0cm 0cm]{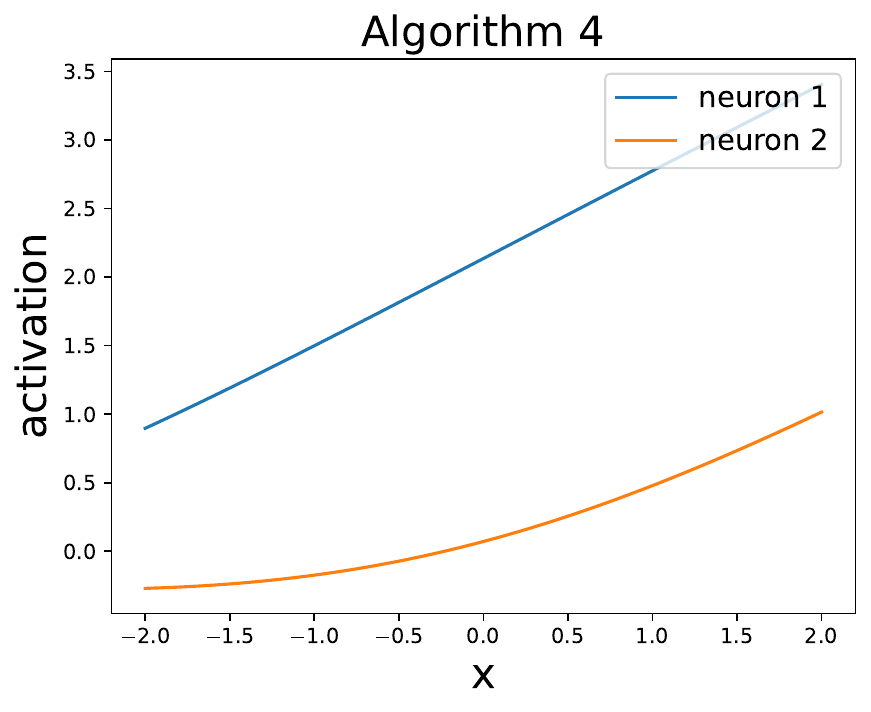}
    \caption{}
    \end{subfigure}
    \caption{Activations of both neurons as functions of inputs, for four algorithms.}
    \label{fig:alg-actx}
\end{figure}

\begin{figure}[h]
    \centering
    \begin{subfigure}[]{0.23\textwidth}
\includegraphics[width=\linewidth, trim=0cm 0cm 0cm 0cm]{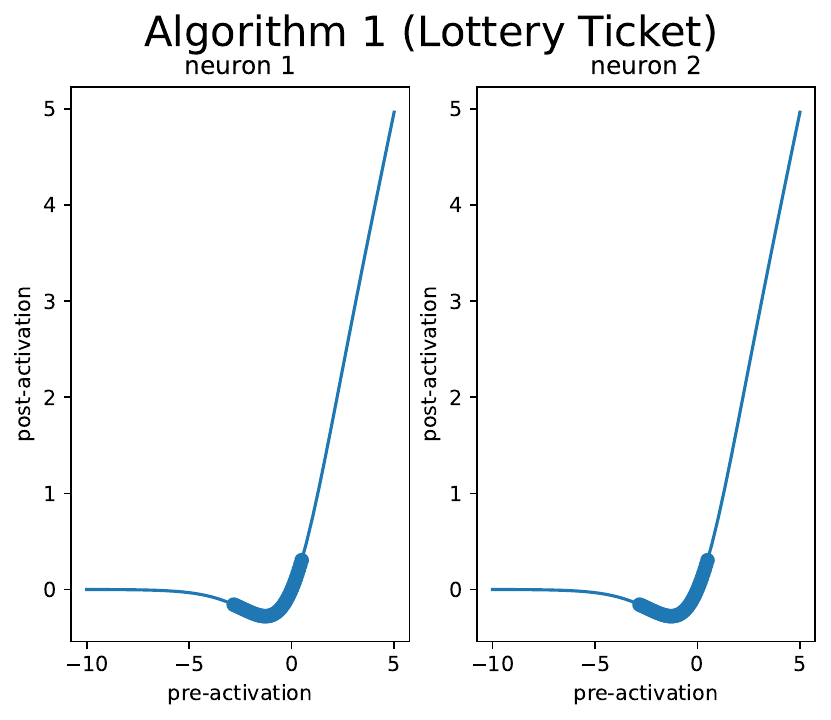}
    \caption{}
    \end{subfigure}
    \begin{subfigure}[]{0.23\textwidth}
    \includegraphics[width=\linewidth, trim=0cm 0cm 0cm 0cm]{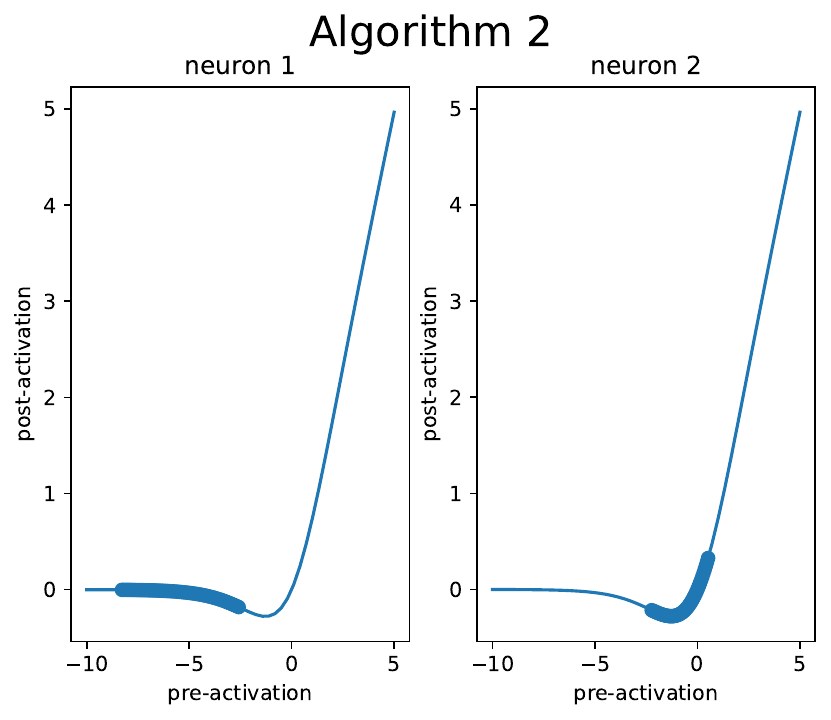}
    \caption{}
    \end{subfigure}
    \begin{subfigure}[]{0.23\textwidth}
    \includegraphics[width=\linewidth, trim=0cm 0cm 0cm 0cm]{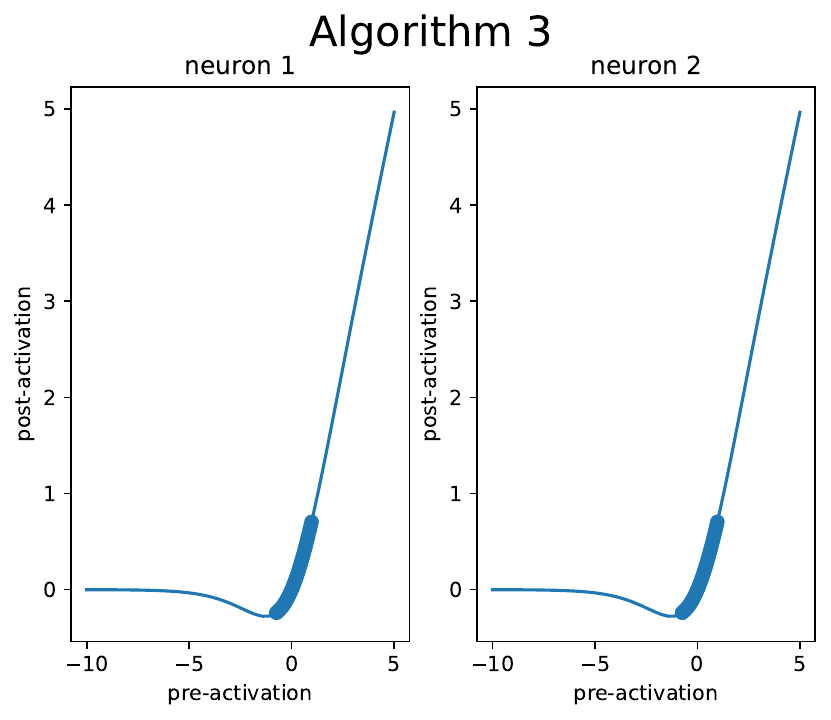}
    \caption{}
    \end{subfigure}
    \begin{subfigure}[]{0.23\textwidth}
    \includegraphics[width=\linewidth, trim=0cm 0cm 0cm 0cm]{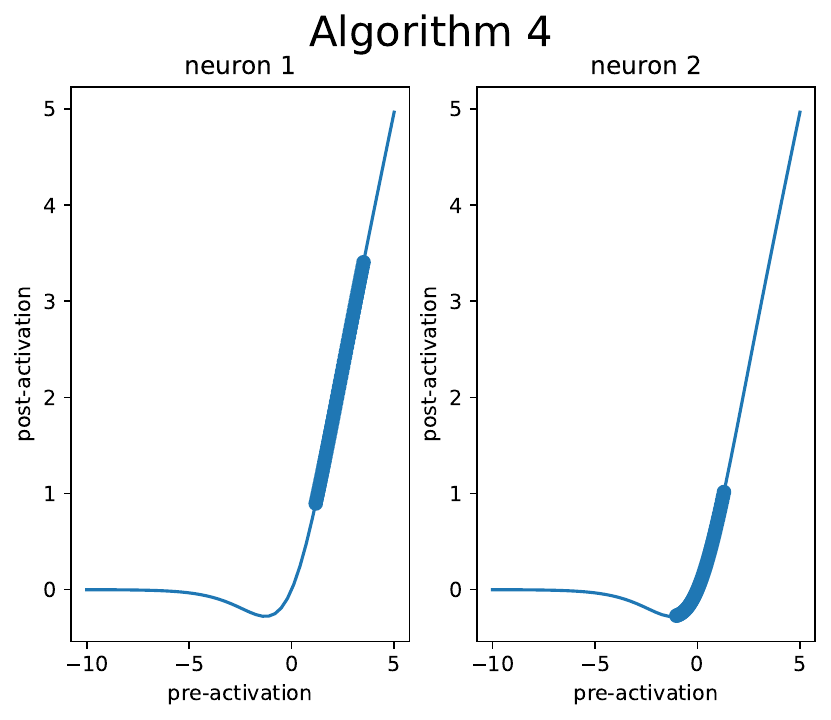}
    \caption{}
    \end{subfigure}
    \caption{Post-activations as functions of pre-activations for both neurons in four algorithms.}
    \label{fig:alg-actpre}
\end{figure}

For networks as small as only having two neurons, what does it even mean that they represent algorithms? We study how pre-activations and post-activations behave for both silu neurons in the range of samples. We show the post-activations as functions of inputs $x$ in Figure~\ref{fig:alg-actx}, and the post-activations as a function of pre-activations in Figure~\ref{fig:alg-actpre}. It is clear from both plots that Algorithm 1 and 3 are symmetric, while Algorithm 1 and 4 are asymmetric. In particular from Figure~\ref{fig:alg-actpre}, four algorithms leverages different parts of the silu networks. Algorithm 1: both neurons are leveraging the intermediate nonlinear part (similar to a quadratic function there). Algorithm 2: one neuron leverages the nagtive saturating part, while the other neuron leverages the intermediate nonlinear part. Algorithm 3: both neurons leverage the quasi-linear part at the boundary between nonlinear and linear parts. Algorithm 4: one neuron is activated at the linear section, while the other neuron stays at the boundary between linear and nonlinear segments.

\section{Disentangling lottery tickets and their correlations}\label{app:disentangle}

\begin{figure}[htbp]
    \centering
    \begin{subfigure}[]{1.0\textwidth}
\includegraphics[width=0.9\linewidth, trim=0cm 0cm 0cm 0cm]{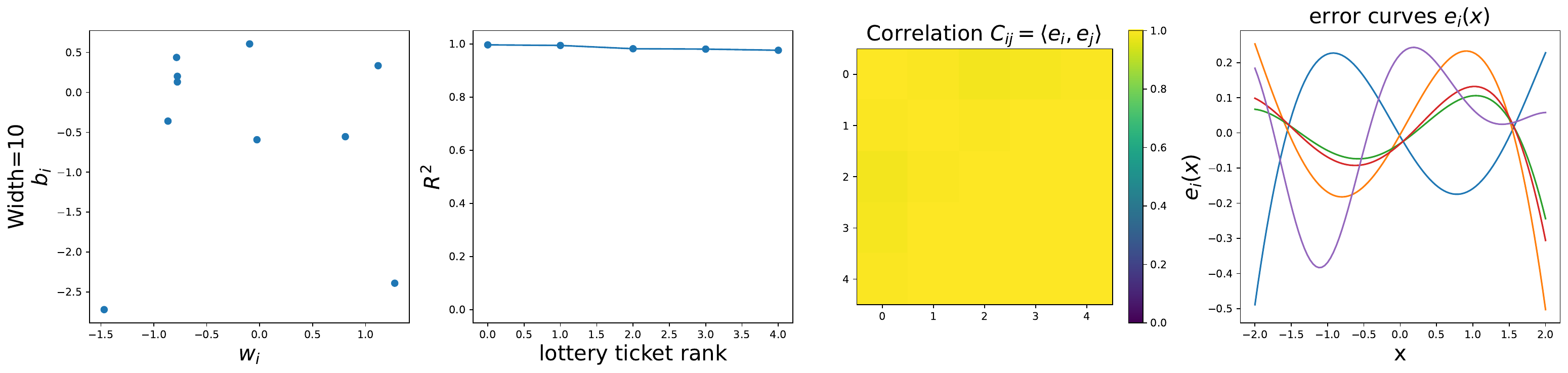}
    \caption{}
    \label{fig:lt-analysis-10}
    \end{subfigure}
    \begin{subfigure}[]{1.0\textwidth}
    \includegraphics[width=0.9\linewidth, trim=0cm 0cm 0cm 0cm]{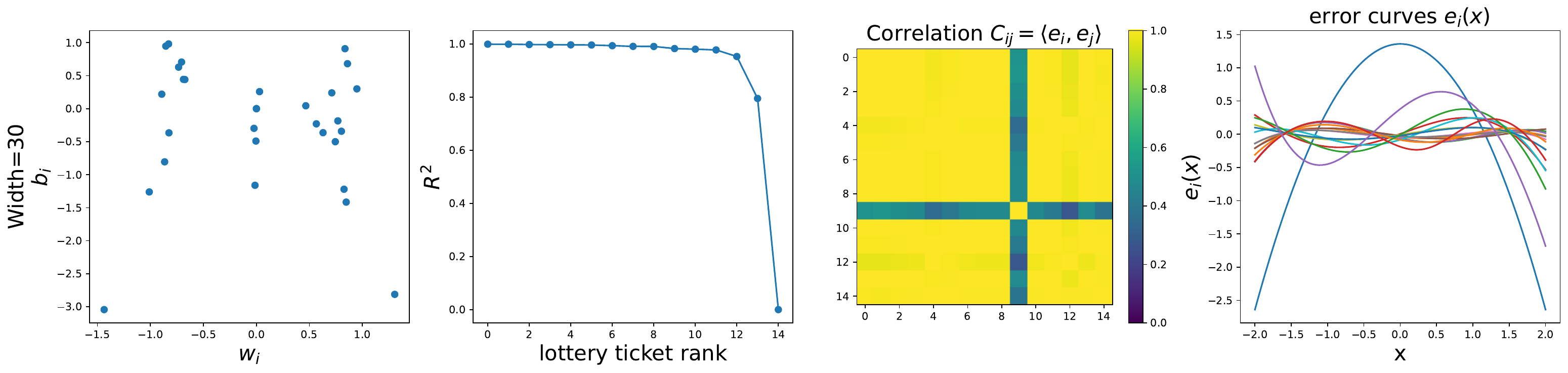}
    \caption{}
    \label{fig:lt-analysis-30}
    \end{subfigure}
    \begin{subfigure}[]{1.0\textwidth}
    \includegraphics[width=0.9\linewidth, trim=0cm 0cm 0cm 0cm]{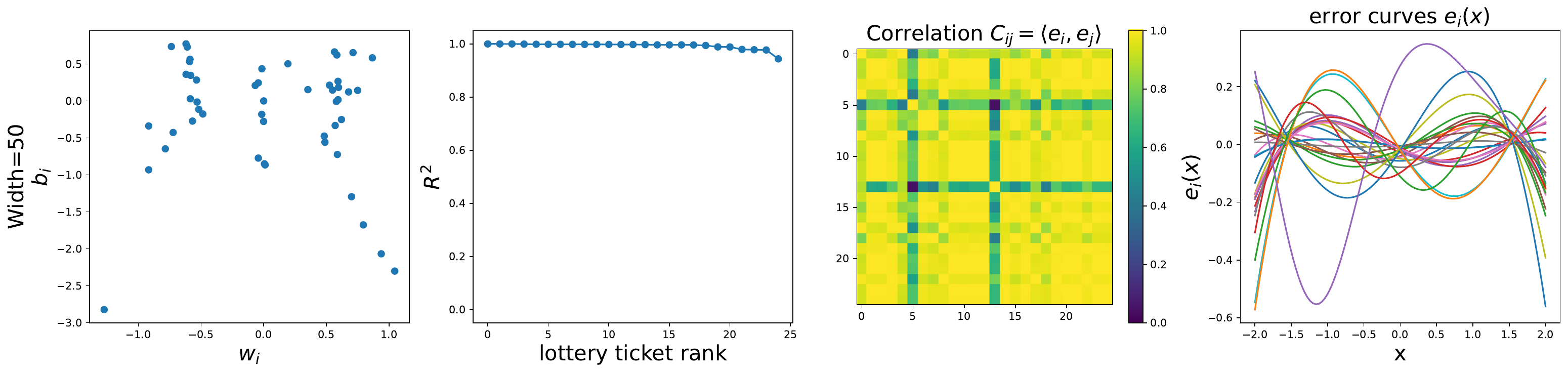}
    \caption{}
    \end{subfigure}
    \begin{subfigure}[]{1.0\textwidth}
    \includegraphics[width=0.9\linewidth, trim=0cm 0cm 0cm 0cm]{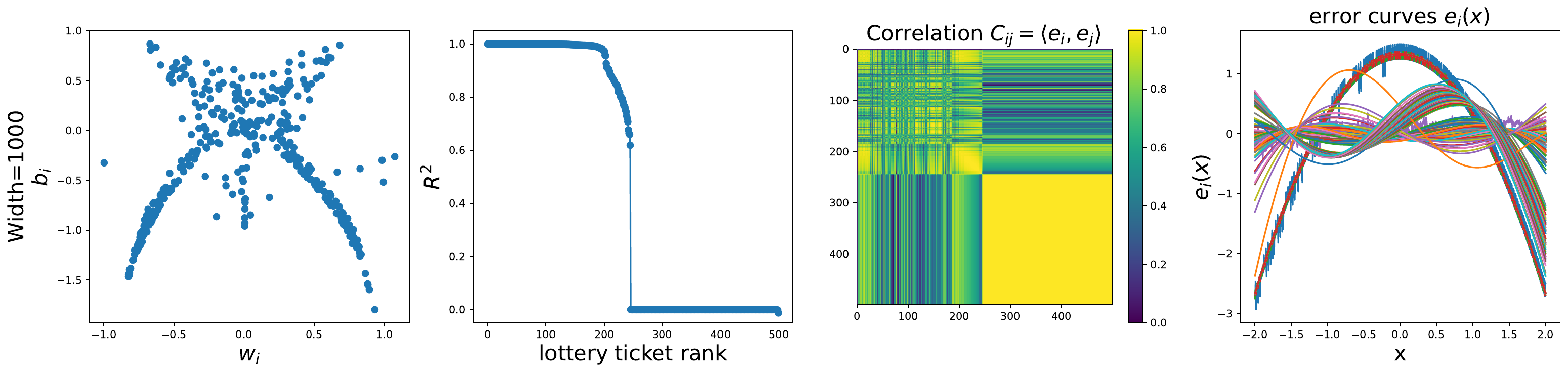}
    \caption{}
    \end{subfigure}

    \caption{Disentangling lottery tickets and measureing their correlation. For top to bottom: width = 10, 30, 50, 1000. For left to right - The first left: weights and biases for the first layer. The second left: $R^2$ (coefficients of determination) of lottery tickets and the target function $f(x)=x^2$. The second right: correlation matrices for lottery tickets, measured by inner products of error functions. The first right: error functions for lottery tickets.}
    \label{fig:distangle-lt}
\end{figure}

In the main paper, we defined lottery tickets in narrow networks, and statistically studied lottery tickets in wider networks. Is it possible to literally disentangle a wide network into many lottery tickets? In general this is difficult since it is unclear what a lottery ticket means for a general task. However, for the toy case $f(x)=x^2$, we find that a lottery ticket consists of two symmetric neurons. Symmetric neurons refer to  a pair of neurons whose incoming weights and biases are $(w,b)$ and $(-w,b)$. This is a criterion we can use to find and disentangle lottery tickets in wide networks: we want to partition $N$ neurons into $N/2$ groups (each group has size 2), such that neurons in the same group are approximately symmetric neurons. In Figure~\ref{fig:distangle-lt} first column, we see that the neurons are symmetrically distributed, hence it makes sense to pair them up. After pairing up neurons into lottey tickets, we can evaluate the quality of each lottery ticket by measuring its correlation coefficient $R^2$ with the target function $f(x)=x^2$ (Figure~\ref{fig:distangle-lt} second column). Almost all lottery tickets have very high quality for $N=10,30,50$, i.e., $R^2\approx 1$. However, the $N=1000$ network only has approximately half of high-quality lottery tickets, while the other half are completely useless. Although there are many high-quality lottery tickets, they are unfortunately highly correlated (Figure~\ref{fig:distangle-lt} third column). For example, $N=10$ seems to only have 1 independent lottery ticket, $N=30$ has 2 independent lottery tickets, and $N=50$ has 3 independent lottery tickets. For each lottery ticket, we can compute its error function $e_i(x)$ as the difference between the best linear predictor and the target function (Figure~\ref{fig:distangle-lt} fourth column). Error functions seem to be highly correlated, e.g., $|e_i(x)|$ tends to be small around $x=0, \pm 1.6$ and large elsewhere. The high correlation among lottery tickets may imply that these networks are unnecessarily redundant. We would like to investigate in the future whether we can use this insights from lottery tickets to reduce these redundancy for network pruning. A preliminary study is carried out below in Appendix~\ref{app:distill}.

\section{Distilling a narrow network from a wide network}\label{app:distill}

\begin{figure}
    \centering
    \includegraphics[width=0.5\linewidth]{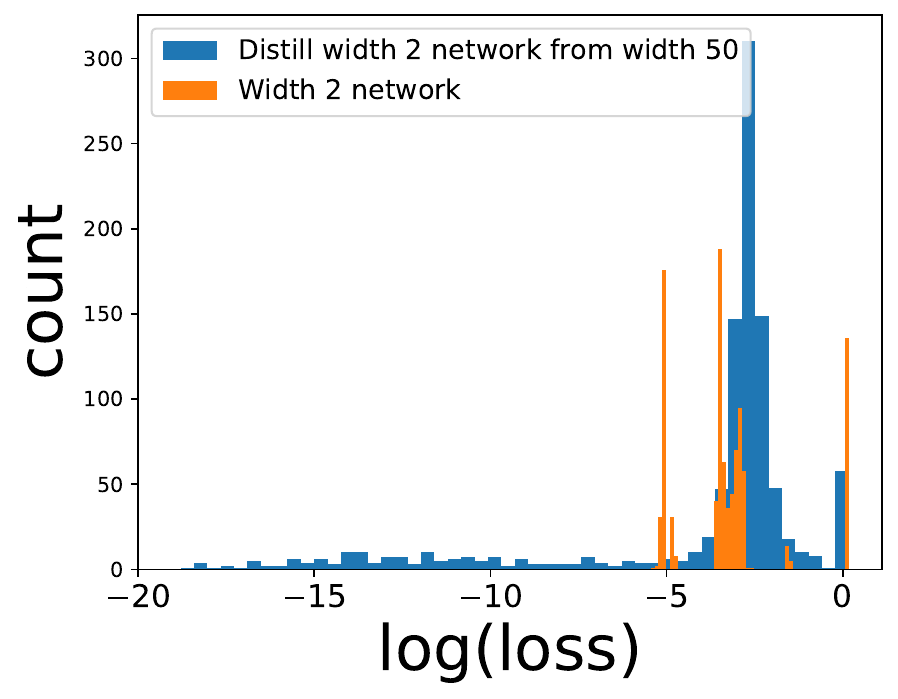}
    \caption{We distill winning lottery tickets (width 2) from width 50 networks. The loss histogram of the distilled width 2 networks is shown in blue, compared to the loss histogram of width 2 networks in orange. On average, the distilled width 2 networks have similar or even worse performance than non-distilled ones, however, there are distilled networks which are extremely good, which achieve losses smaller than machine precision $<10^{-16}$.}
    \label{fig:hist-distill-lt}
\end{figure}

The techniques in Appendix~\ref{app:disentangle}, i.e., disentangling a wide network into lottery tickets, allows us to distill a much smaller network from a wide network. Concretely, we take a width $N=50$ networks (25 lottery tickets) and select out the best lottery ticket (the one with highest $R^2$ with targets). Now we use this lottery ticket to construct a width 2 network: weights and biases of the first layer are obtained by directly copying the lottery ticket, while the weights and biases of the second layer are obtained via linear regression. We call them distilled networks. We do this distillation for 1000 random seeds, and plot their losses in Figure~\ref{fig:hist-distill-lt} (blue). We also compare to non-distilled width 2 networks (orange). On average, the distilled width 2 networks have similar or even worse performance than non-distilled ones, however, there are distilled networks which are extremely good, which achieve losses smaller than machine precision $<10^{-16}$. These distilled network are indeed "lottery tickets of lottery tickets". However, it seems impossible to fine-tune these lottery tickets further, since they have extremely ill-conditioned weights. For the distilled network with lowest loss ($1.6\times 10^{-19}$), its parameters are:
\begin{equation}
    \mat{w} = (-5.9\times 10^{-8}, 1.2\times 10^{-5}), \mat{b} = (-2.8\times 10^{-7},  7.6\times 10^{-5}), \mat{v} = (5.5\times 10^{12}, 2.6\times 10^{10}), c = -2.2\times 10^5.
\end{equation}

\end{document}